\documentclass{article}

\usepackage{microtype}
\usepackage{graphicx}
\usepackage{subcaption}
\usepackage{booktabs} 

\usepackage{hyperref}

\usepackage[preprint]{icml2026}

\usepackage{amsmath}
\usepackage{amssymb}
\usepackage{mathtools}
\usepackage{amsthm}

\usepackage[capitalize,noabbrev]{cleveref}

\theoremstyle{plain}

\theoremstyle{definition}

\theoremstyle{remark}

\usepackage[textsize=tiny]{todonotes}

\usepackage{xspace}
\usepackage[table]{xcolor}
\usepackage{booktabs}
\newcommand{\method}{PreFlect\xspace}

\usepackage{booktabs}
\usepackage{multirow} 
\usepackage{array} 
\usepackage{algorithm}
\usepackage{algorithmic}
\usepackage{amsmath}

\usepackage{wasysym}
\usepackage{hyperref}
\usepackage{url}

\usepackage{tcolorbox}
\usepackage{fontawesome}
\tcbuselibrary{skins, breakable}
\usepackage{listings}
\usepackage[inline]{enumitem}
\usepackage{makecell}
\newcommand{\resultitem}[2]{%
    \item[\small\bfseries\itshape #1] {\small #2}%
}

\newcommand{\ifcomments}{\iftrue}

\icmltitlerunning{PreFlect: From Retrospective to Prospective Reflection in Large Language Model Agents}

\begin{document}

\twocolumn[
  \icmltitle{PreFlect: From Retrospective to Prospective Reflection \\
  in Large Language Model Agents}



  \icmlsetsymbol{equal}{*}

  \begin{icmlauthorlist}
    \icmlauthor{Hanyu Wang}{psu}
    \icmlauthor{Yuanpu Cao}{psu}
    \icmlauthor{Lu Lin}{psu}
    \icmlauthor{Jinghui Chen}{psu}
  \end{icmlauthorlist}

  \icmlaffiliation{psu}{College of Information Sciences and Technology, The Pennsylvania State University, State College, PA, USA}

\icmlcorrespondingauthor{Hanyu Wang}{hbw5365@psu.edu}
\icmlcorrespondingauthor{Jinghui Chen}{jzc5917@psu.edu}

  \icmlkeywords{Machine Learning, ICML}

  \vskip 0.3in
]



\printAffiliationsAndNotice{}  

\begin{abstract}
  Advanced large language model agents typically adopt self-reflection for improving performance, where agents iteratively analyze past actions to correct errors. However, existing reflective approaches are inherently retrospective: agents act, observe failure, and only then attempt to recover. In this work, we introduce \textbf{\method}, a prospective reflection mechanism that shifts the paradigm from post hoc correction to pre-execution foresight by criticizing and refining agent plans before execution. To support grounded prospective reflection, we distill planning errors from historical agent trajectories, capturing recurring success and failure patterns observed across past executions. Furthermore, we complement prospective reflection with a dynamic re-planning mechanism that provides execution-time plan update in case the original plan encounters unexpected deviation. Evaluations on different benchmarks demonstrate that \method significantly improves overall agent utility on complex real-world tasks, outperforming strong reflection-based baselines and several more complex agent architectures. Code will be updated at \url{https://github.com/wwwhy725/PreFlect}.
\end{abstract}

\section{Introduction}
\label{sec:intro}

Large language model agents \citep{wu2024autogen, hu2025owl, li2025search}, empowered by advanced LLMs \citep{guo2025deepseek, comanici2025gemini, yang2025qwen3}, have demonstrated competence in handling various real-world tasks. However, when faced with tasks requiring highly complex reasoning and multiple tools handling, even a minor error may ruin all the efforts \citep{press2023measuring}. To overcome the fragility of single-pass reasoning, recent research has increasingly integrated self-reflection mechanisms into agent architecture \citep{gou2023critic, shinn2023reflexion, zhou2023language}. Unlike standard CoT approaches, reflection enables agents to act as their own critics: scrutinizing past execution trajectories, diagnosing action-level errors, and iteratively refining their future strategies \citep{madaan2023self}.

\begin{figure}[t]
    \centering
    \includegraphics[width=0.95\linewidth]{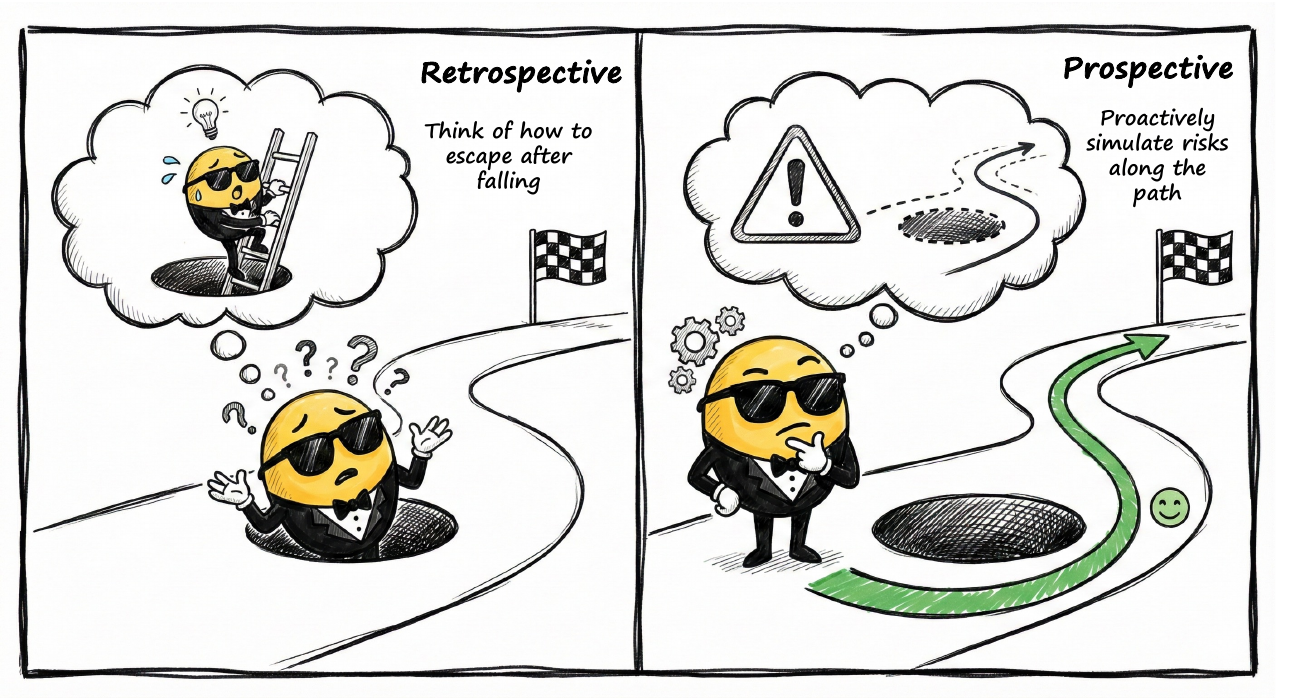}
    \caption{\textbf{Retrospective vs. Prospective Reflection.} (Left) The retrospective agent triggers reflection only after encountering a failure. (Right) The prospective agent anticipates potential risks before execution, allowing it to adjust its plan and successfully bypass the obstacle.}
    \vspace{-15pt}
    \label{fig:mot}
\end{figure}

Despite these advances, most existing reflection paradigms remain fundamentally \textbf{retrospective}, operating in a reactive manner that triggers correction only after failures have already occurred. While effective in some settings, such post-hoc reflection faces several inherent limitations in complex environments. Specifically, retrospective correction is often inadequate when actions produce \textbf{irreversible consequences}, such as accidentally deleting an important file. Additionally, diagnosing errors after execution can introduce substantial \textbf{trajectory-level noise}: by storing both failed attempts and subsequent fixes in memory, agents may suffer from contextual interference that destabilizes future decision-making. Furthermore, retrospective methods typically rely on repeated trial-and-error cycles, leading to significant computational cost and increased token latency as agents iteratively recover from mistakes.

To address these limitations, we propose \textbf{P}rospective \textbf{Reflect}ion (\textbf{\method}), a novel framework that shifts the reflection paradigm from reactive correction to proactive prevention, equipping agents with a prospective capability to anticipate potential pitfalls.
Central to this shift is identifying when reflection should intervene in the agent’s decision process. Existing methods typically reflect only after actions have been executed, when errors may already be costly or irreversible. In contrast, we observe that the planning stage provides a critical opportunity for proactive control: it is the point at which the agent commits to a strategy, yet has not begun to act on it.
Building on this insight, we introduce a planning-phase reflection mechanism that operates in the cognitive window between plan generation and action execution. By evaluating and revising plans before they are carried out, \method decouples error detection from irreversible outcomes, allowing agents to scrutinize their strategies for potential risks prior to execution.

However, the efficacy of prospective reflection is often constrained by the inherent difficulty of anticipating future failures. In the absence of structured guidance, aimless critique may even introduce more risks, such as hallucinations \citep{zhu2025llm}. To resolve this issue, we incorporate the ``Planning Errors'' into \method, which serves as an experiential anchor by distilling recurrent failure and success patterns from past agent trajectories. Specifically, upon generating a plan, the agent cross-references its proposed strategies against the planning errors to diagnose specific vulnerabilities such as hazardous tool usage. This guided reflection allows the agent to transform abstract foresight into concrete diagnosis by identifying semantic similarities between the current plan and historical pitfalls. Consequently, the agent can proactively substitute risky components with reliable alternatives, resulting in a robust and optimal trajectory that is empirically grounded before any action is committed to the environment.

Moreover, although the planning errors provide reliable and grounded references, proactively reflecting on future errors may still be insufficient under the uncertainty of complex environments. Relying solely on pre-execution critique can leave blind spots during execution and lead to oversights. Therefore, we incorporate a dynamic re-planning mechanism to complement planning-phase reflection by adaptively updating the agent’s plan during execution.  Unlike traditional architectures that adhere to rigid, fixed interval planning cycles, the dynamic re-planning continuously monitors the execution state to evaluate whether the current trajectory remains viable or necessitates a strategic shift. For instance, if the agent originally plans to search for a certain query while repeated failures, indicating infeasibility, the agent can trigger re-planning to revise its strategy and pivot toward alternative approaches.

Together, prospective reflection and dynamic re-planning form a unified framework that supports both proactive planning and robust execution.
To validate \method, we conduct comprehensive experiments across multiple benchmarks. Results show that \method achieves consistent performance improvements while maintaining reasonable token overhead. To sum up, our contributions are threefold:

\vspace{-15pt}
\begin{itemize}
\setlength{\itemsep}{-2pt}
    \item We propose \method, a prospective reflection framework that enables LLM agents to proactively identify and avoid critical failures before execution.
    \item To further improve the effectiveness and efficiency of \method, we introduce two key components: the planning errors and  dynamic re-planning mechanism, which respectively provide experiential priors to assist reflection and adaptively revise plans during execution.
    \item Extensive experiments demonstrate that \method significantly outperforms existing reflection-based baselines and competitive agent frameworks. Furthermore, transferability results show that \method generalizes across different agent architectures.
\end{itemize}
\vspace{-15pt}

\begin{figure*}[t]

    \centering \includegraphics[width=0.95\linewidth]{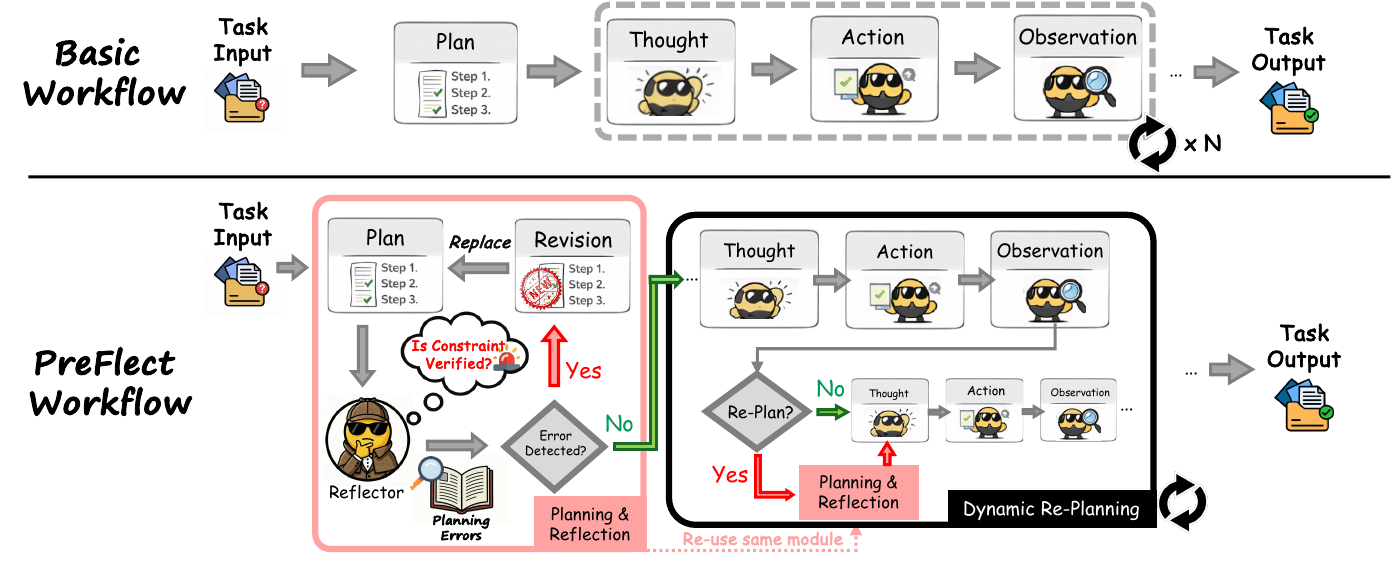}
    \caption{The architecture of \textbf{\method} comparing with basic agent workflow (top). \method integrates a prospective reflection loop (pink box) into the planning stage. The bottom-right panel illustrates dynamic re-planning workflow.}
    \vspace{-10pt}
    \label{fig:main}
\end{figure*}

\section{Related Work}
\label{sec:related_work}

\noindent\textbf{LLM Agents.} LLM-based agents enhance reasoning by integrating external modules for planning, memory, and tool use. The ReAct \citep{yao2022react} architecture established the iterative ``reason-act'' paradigm, which has since evolved into specialized research on test-time searching \citep{yao2023tree, besta2024graph}, tool tuning \citep{zeng2025tool, qian2025toolrl, jiang2025verltool}, and memory structuring \citep{xu2025mem, chhikara2025mem0}. Beyond modular improvements, recent frameworks introduce diverse coordination strategies: AutoAgent \citep{chen2023autoagents} and OWL \citep{hu2025owl} utilize dynamic multi-agent coordination and role-playing for task decomposition, while Alita \citep{qiu2025alita} enables autonomous expansion via refined task-specific MCPs. Finally, Smolagents \citep{smolagents} prioritizes efficiency through lightweight, code-based actions and local deployment integration.

\noindent\textbf{Self-Reflection.} Self-reflection enables LLM agents to autonomously critique past executions and iteratively refine subsequent strategies. Reflexion \citep{shinn2023reflexion} established the foundation for this paradigm by introducing verbal reinforcement to store linguistic feedback to prevent recurring errors. Self-Refine \citep{madaan2023self} employs an iterative loop where the model generates, critiques, and corrects its own output without external supervision. Recent advancements explore more granular reflection strategies: AR \citep{wang2024devil} focuses on real-time adaptation by preparing backup moves and switching to alternatives upon primary execution failure. MIRROR \citep{guo2025mirror} further extends this by integrating inter-agent communication with intra-agent self-critique. SAMULE \citep{ge2025samule} shifts from prompting to specialized training, fine-tuning a reflection LLM to provide high-quality feedback to an executor agent. 

\section{Methodology}
\label{sec:method}

We organize this section as follows: Section~\ref{subsec:overview} provides an overview of our proposed method. In Section~\ref{subsec:reflect}, we expand on the core concept of prospective reflection, with Section~\ref{subsubsec:plan_error} introducing planning errors and Section~\ref {subsubsec:ref_rev} explaining how \method reflects on and revises the plan. Finally, in Section~\ref{subsec:arp}, we show how dynamic re-planning and prospective reflection complement each other.

\subsection{Overview of \method}
\label{subsec:overview}

Figure~\ref{fig:main} provides an overview of \method by contrasting it with a standard agent workflow and illustrating how prospective reflection and execution-time dynamic re-planning are integrated into a unified loop.
The top of the figure shows the \emph{basic agent workflow}. Given a task input, the agent first generates a plan that decomposes the task into a sequence of steps based on the currently known conditions. The agent then executes the plan through an iterative \textit{think--act--observe} loop, during which it reasons about the next step, takes an action, and observes the environment. This execution loop repeats until the task is completed, at which point the final output is produced.
The bottom of the figure depicts the workflow of \method. The left part highlights the \emph{planning and prospective reflection module}. After an initial plan is generated, a reflector examines the plan \emph{before execution} by leveraging a set of distilled planning errors, which summarize common failure and success patterns observed from historical agent trajectories. The reflector detects potential planning errors and revises the plan accordingly. This process iterates until a validated plan is produced, which is then passed to the execution stage.
The right part of the figure shows how execution proceeds under \method and how \emph{execution-time dynamic re-planning} is handled. Similar to the basic workflow, the agent executes the validated plan using the \textit{think--act--observe} loop. During execution, the agent continuously monitors the trajectory. When the current trajectory becomes infeasible or deviates from the validated plan, a re-planning trigger is activated. Importantly, dynamic re-planning does not rewind or discard the past trajectory; instead, it \emph{appends a new planning and prospective reflection phase} that is conditioned on the execution history so far. As illustrated inside the dynamic re-planning workflow, this re-planning step reuses the same planning and prospective reflection module shown on the left, ensuring that all newly generated plans, whether initial or re-generated during execution, are validated prospectively before further actions are taken.

Overall, \method forms a single closed-loop system in which prospective reflection is applied both prior to initial execution and whenever re-planning is required, enabling the agent to anticipate and mitigate planning errors before they manifest as irreversible execution failures. In the following, we will step into the design details of each component.

\subsection{Prospective Reflection}
\label{subsec:reflect}

The core component of our proposed \method is a \textit{prospective reflection} mechanism where the agent reflects on the plan based on distilled error patterns, identifies critical errors, and replaces the original plan with a revised one to avoid future failures.

\subsubsection{Planning Errors}
\label{subsubsec:plan_error}
Unlike retrospective reflection, which corrects errors through post-execution trajectory review, prospective reflection must reason about potential failures before they occur, and is therefore subject to greater uncertainty. Without grounded guidance, such forward-looking critique can become unreliable or overly speculative. To address this challenge, we adopt an \emph{offline} distillation process that extracts common error modes and success/failure patterns from an agent’s past experience, providing experiential priors that enable more accurate prospective reflection.

\noindent\textbf{Trajectory Collection.} 
Constructing Planning Errors requires leveraging the agent’s own experience across diverse tasks. For each task, we sample three trajectories and identify cases with \textit{mixed outcomes}, where the agent produces both successful and failed attempts. This contrastive setting highlights the key differences between effective and ineffective strategies, facilitating the extraction of recurring planning-level failure patterns. To promote generalization and avoid overfitting, the data used for Planning Error construction is disjoint from all evaluation benchmarks.

\noindent\textbf{Diagnosis.} 
Given these mixed trajectories, we prompt an LLM to conduct a comparative diagnostic analysis. The diagnosis focuses on identifying critical errors in failed trajectories that arise from planning deficiencies and could be avoided through improved planning. In parallel, it analyzes how successful trajectories circumvent or resolve these pitfalls. Each diagnostic result produces one or more error types, along with corresponding descriptions, impacts, and supporting evidence. This process allows us to isolate the exact planning failures that differentiate success from failure.

\vspace{-5pt}
\begin{tcolorbox}[
    colback=gray!5,
    colframe=gray!50!black,
    title=\textbf{An Example in Planning Errors},
    fonttitle=\bfseries,
    breakable
]

\textbf{Error Type:} \texttt{\textcolor{red!70!black}{insufficient constraint verification}}

\tcbline
\vspace{-8pt}
\begin{description}[
    leftmargin=0.0em,      
    font=\bfseries\itshape, 
    itemsep=-0.5em    
]
    
    \resultitem{Description:}{The plan identifies a potential answer but fails to rigorously verify it against all specific constraints or conditions in the prompt \dots It relies on \dots without including steps to explicitly confirm that the candidate satisfies every requirement \dots}
    
    \resultitem{Success Example:}{In a task identifying an actor \dots the agent identified Michael Gambon and correctly verified that \dots ensuring he met the specific constraint.}
    
    \resultitem{Failure Example:}{In a task identifying an actor \dots the agent identified Richard Griffiths because \dots However, the plan failed to verify \dots leading to an incorrect answer.}

\end{description}

\end{tcolorbox}

\noindent\textbf{Aggregation.} The final stage of our framework synthesizes individual diagnoses into a unified taxonomy of planning errors. We employ an LLM-based aggregator to construct this taxonomy by iteratively comparing each new diagnostic entry against the existing error set. For each entry, the aggregator determines whether to instantiate a new error category, merge it into an existing profile, or discard it as redundant. This process distills generalizable planning-level failure modes that transcend specific task instances, ensuring broad applicability. In parallel, the LLM summarizes empirical evidence and trajectory impacts into concrete success and failure patterns. The resulting Planning Errors therefore connect high-level reasoning deficiencies with recurring behavioral signatures, providing informative priors for downstream reflection. After the initial LLM-driven synthesis, we perform a final manual refinement step to remove categories that are overly narrow or task-dependent (e.g., errors specific to a particular query formulation). This yields a finalized set of three core error types:
\begin{enumerate*}[label=\arabic*)]
    \item insufficient constraint verification
    \item ineffective tool selection
    \item shallow content verification,
\end{enumerate*}
capturing failures in constraint maintenance, tool usage, and content understanding. Notably, these errors are designed to be domain-agnostic, maximizing their utility across diverse agentic tasks. For more details, please refer to Appendix~\ref{appendix:error}.


\subsubsection{Reflection \& Revision}
\label{subsubsec:ref_rev}
\noindent\textbf{Error Identification.} After a plan is generated, a reflector agent evaluates it to identify potential critical errors before execution. To improve accuracy and reduce false positives, the reflector leverages the distilled Planning Errors as grounded reference priors. By conditioning reflection directly on these domain-agnostic failure modes, the agent can perform more structured and reliable prospective critique.

Particularly, the agent summarizes its previous actions to have a comprehensive understanding of the task and current states. Meanwhile, the agent analyzes available tools and learns how and why some tools fail or succeed in the environment of the current task.
This self-awareness helps foster more accurate and fact-based prospective reflection, enabling foresight into future scenarios, such as knowing whether a tool should be avoided under a certain circumstance before even attempting it. After gathering enough information about the current states, the reflector critiques the proposed plan by checking whether it exhibits patterns aligned with any known Planning Errors. If a flaw is detected, the agent revises the plan accordingly, producing a higher-quality strategy prior to action execution.

\noindent\textbf{Plan Revision.} Given the identified errors in the previous step, the agent now switches its attention to how to comprehensively develop an updated plan to impact the future positively. Notably, the planning errors collection contains contrastive examples of short trajectory segments where positive examples demonstrate how the agent avoids such an error, while negative ones expand on how the agent is misled to failure. Combining the high-level error description with these concrete cases, the agent is enabled to match its current state with the error information and find out the optimal path toward success.

\subsection{Dynamic Re-Planning}
\label{subsec:arp}
Although Planning Errors provide grounded experiential priors for prospective reflection, reasoning about future failures remains inherently uncertain, and the agent may still overlook critical risks during execution. To complement planning-phase reflection, we introduce a dynamic re-planning mechanism that enables the agent to adapt its strategy online when the environment reveals unexpected obstacles. Notably, whenever re-planning is triggered, prospective reflection is invoked again to assess potential risks in the updated plan.

\noindent\textbf{Dynamic Plan Updating.} 
During execution, the agent continuously gathers new observations through interaction with the environment. Such feedback may invalidate earlier assumptions, causing the original plan to become suboptimal or infeasible. Since many execution-time constraints are only revealed at runtime (e.g., unavailable tool outputs or missing external information), proactive reflection alone cannot guarantee optimal trajectories under evolving states.
To address this, we equip the agent with a \texttt{re-plan} operation that can be triggered whenever progress stalls or feasibility conditions are violated. Following the ReAct paradigm, the agent explicitly reasons about the current execution state and identifies why the existing plan is no longer effective before invoking re-planning. The agent then generates an updated strategy and immediately resumes execution under the revised plan. Implementation details are provided in Appendix~\ref{appendix:imp}.

\section{Experiments}
\label{sec:experiment}

\begin{table*}[ht]
\caption{
    Main experimental results on GAIA and SimpleQA benchmarks. \method is built upon Smolagents framework. All the results are using the pass@1 metric. The best results are shown in \textbf{bold}.
  }
  \label{tab:main}
  \centering
  \resizebox{0.95\textwidth}{!}{
      \begin{tabular}{llcccccccc}
        \toprule
          \multirow{2}{*}{\makecell[l]{\textbf{Agent} \\ \textbf{Framework}}} & \multirow{2}{*}{\makecell[l]{\textbf{Reflective} \\ \textbf{Method}}} & \multicolumn{4}{c}{\textbf{GAIA (\%)}} & \multicolumn{4}{c}{\textbf{SimpleQA (\%)}} \\
        \cmidrule(lr){3-6}  \cmidrule(lr){7-10}
       & & Level 1 ($\uparrow$) & Level 2 ($\uparrow$) & Level 3 ($\uparrow$) & Total ($\uparrow$) & Corr. ($\uparrow$) & Incor. ($\downarrow$) & N.A. ($\downarrow$) & C./Att. ($\uparrow$) \\
        \midrule
        \rowcolor{gray!15}
        \multicolumn{10}{c}{\textit{Backbone LLM: GPT-4.1}} \\
        ReAct & - & 41.51 & 33.72 & 7.69 & 32.12 & 61 & 19 & 20 & 76.25 \\
        ReAct & Reflexion & 49.06 & 34.88 & 15.38 & 36.36 & 71 & 21 & 8 & 77.17 \\
        ReAct & Self-Refine & 39.62 & 39.53 & 11.54 & 35.15 & 74 & 18 & 8 & 80.43 \\
        Smolagents & - & 56.60 & 47.67 & 19.23 & 46.06 & 72 & 16 & 12 & 81.82 \\
        Smolagents & Reflexion & 56.60 & 54.65 & 19.23 & 49.70 & 79 & 20 & \textbf{1} & 79.80 \\
        Smolagents & Self-Refine & 58.49 & 52.33 & 23.08 & 49.70 & 78 & 18 & 4 & 81.25 \\
        {Smolagents} & \textbf{\method (ours)} & \textbf{71.70} & \textbf{55.81} & \textbf{38.46} & \textbf{58.18} & \textbf{83} & 15 & 2 & \textbf{84.69} \\
        \midrule
        \rowcolor{gray!15}
        \multicolumn{10}{c}{\textit{Backbone LLM: Gemini-2.5-pro}} \\
        ReAct & - & 56.60 & 45.35 & 30.77 & 46.67 & 55 & 16 & 29 & 77.46 \\
        ReAct & Reflexion & 56.60 & 47.67 & 30.77 & 47.88 & 70 & 18 & 12 & 79.55 \\
        ReAct & Self-Refine & 56.60 & 44.19 & 34.62 & 46.67 & 62 & 24 & 14 & 72.09 \\
        Smolagents & - & 54.72 & 52.32 & 34.62 & 50.30 & 76 & 18 & 6 & 80.85 \\
        Smolagents & Reflexion & 60.38 & 52.33 & \textbf{38.46} & 52.73 & 78 & 22 & \textbf{0} & 78.00 \\
        Smolagents & Self-Refine & 60.38 & 50.00 & 34.62 & 50.91 & \textbf{81} & 17 & 2 & 82.65  \\
        {Smolagents} & \textbf{\method (ours)} & \textbf{67.92} & \textbf{60.47} & \textbf{38.46} & \textbf{59.39} & \textbf{81} & \textbf{13} & 6 & \textbf{86.17}  \\
        \bottomrule
      \end{tabular}
     }
\end{table*}

\subsection{Experiment Settings}

\noindent\textbf{Datasets.} To evaluate the general task completion performance of \method, we conduct comprehensive experiments on the following datasets with complete agentic settings. 
\begin{enumerate*}[label=\arabic*)]
    \item \textbf{GAIA} \citep{mialon2023gaia} is a general AI assistant benchmark that evaluates AI agent's ability to complete real-world tasks, covering multi-step reasoning, fact checking, web browsing, and tool usage. We select the validation set.
    \item \textbf{SimpleQA} \citep{wei2024measuring} is a challenging benchmark that evaluates the ability to answer fact-seeking questions. Due to resource limitations, we randomly select 100 samples from the test set.
\end{enumerate*} 
Note that for constructing the planning errors, we leverage \textbf{HotpotQA} \citep{yang2018hotpotqa} and \textbf{MuSiQue} \citep{trivedi2022musique} datasets, ensuring no overlap with the test benchmarks while challenging the agent's diverse capabilities.

\noindent\textbf{Baselines.} We compare with various baselines to validate the effectiveness of our proposed method. In particular, we select Reflexion \citep{shinn2023reflexion} and Self-Refine \citep{madaan2023self}, which are standard retrospective reflection methods. We omit \citet{wang2024devil} as it targets step-level anticipation and is not directly applicable to GAIA-style long-horizon planning tasks. Also, we compare with several agent frameworks, such as ReAct \citep{yao2022react}, Smolagents and its Open Deep Research (ODR) system \citep{smolagents}, AutoAgent \citep{chen2023autoagents}, Magnetic-1 \citep{fourney2024magentic}, HAL \citep{hal}, OWL \citep{hu2025owl}, and OAgent \citep{zhu2025oagents}.

\noindent\textbf{Evaluation Metrics.} For GAIA, we use standard ``pass@1'' which refers to the percentage of problems solved correctly when 1 candidate solution is sampled per task. For SimpleQA, we follow the standard evaluation to report \textbf{Correct} (\textbf{Corr.}), \textbf{Incorrect} (\textbf{Incor.}), \textbf{Not Attempted} (\textbf{N.A.}), and \textbf{Correct Given Attempted} (\textbf{C./Att.}), which refer to the number of correct, incorrect, not attempted, and correct given attempted answers out of all the answers, respectively.

\noindent\textbf{Agent Implementations.} \method (built upon Smolagents) operates under a maximum budget of 20 execution steps and is equipped with standard web search and webpage navigation tools. To satisfy the multimodal tool requirements of GAIA tasks (e.g., involving images or audio), we additionally provide a set of lightweight modality-specific inspection tools. 
Importantly, all compared methods are evaluated under the same tool access and execution constraints to ensure a fair comparison in our main experiments. Further implementation details are deferred to Appendix~\ref{appendix:imp}.

\definecolor{box-blue}{HTML}{8E8BFE}
\definecolor{box-red}{HTML}{FEA3A2}
\definecolor{box-gray}{HTML}{F2F2F2}
\definecolor{error-tag}{HTML}{D32F2F}

\begin{figure*}[ht]
\centering
\begin{tcolorbox}[colback=white, colframe=gray!50, arc=2mm, boxrule=0.5pt, title=\textbf{Qualitative Analysis: An Example From GAIA Level 3 Task}]
    \begin{minipage}[t]{0.45\textwidth}
        \small \textbf{\faSearch \ Task:} \\
        \small Eva Draconis has a personal website which can be accessed on her YouTube page. What is the meaning of the only symbol seen in the top banner that has a curved line that isn't a circle or a portion of a circle? Answer without punctuation.
    \end{minipage}
    \hfill
    \vrule \hfill 
    \begin{minipage}[t]{0.48\textwidth}
        \small \textbf{\faMapSigns \ Ground Truth Trajectory:} \\
        \footnotesize
        \begin{enumerate*}[label=\arabic*., itemjoin={\ $\rightarrow$\ }] 
            \item Google Eva Draconis YouTube \item Find her website URL in her about section \item Enter to identify symbols and their meanings \item Find the only symbol that matches all the requirements \item Note the meaning ``War is not here, this is a land of peace.''
        \end{enumerate*}
    \end{minipage}
    \small
    \begin{tcolorbox}[colback=box-gray, colframe=gray!40, arc=1mm, boxrule=0.5pt, title=\textbf{\faHistory \ Step 1--14: Plans, Actions, and Failure}]
        Agent initiated a multi-step search strategy: scraping YouTube ``About'' sections, scanning social media (Facebook/Patreon), and querying Google Images. \\
        \textbf{Outcome:} All attempts resulted in 404/403 errors or irrelevant results (e.g., Space.com forums), as the primary site (\textit{orionmindproject.com}) was defunct.
    \end{tcolorbox}

    \begin{tcolorbox}[colback=box-blue!10, colframe=box-blue, arc=1mm, boxrule=1pt, title=\textbf{\faLightbulbO \ Dynamic Re-Planning (Step 15)}]
        \textbf{Thought:} All searches for \dots for orionmindproject.com and Eva Draconis site are futile \dots A new approach is mandatory. \\
        \textbf{Re-plan Proposal:} The agent proposed to double-down on Wayback Machine visual captures, attempting to retrieve and process the banner image via the \texttt{inspector} tool.
    \end{tcolorbox}

    \begin{tcolorbox}[colback=box-red!10, colframe=box-red, arc=1mm, boxrule=1pt, title=\textbf{\faCheckSquareO \ Plan Reflection \& Revision}]
        \textbf{Reflection Analysis:} \textcolor{error-tag}{\textbf{[ineffective\_tool\_selection]}} \\
        The plan \dots to locate and process a banner image \dots but \dots visit\_webpage tool cannot retrieve images, and the inspector tools require a publicly accessible image URL \dots The plan does not include a strategy to overcome this technical constraint. \\
        \textbf{Revised Plan:} \textbf{Pivot from visual retrieval to \textit{source-code} excavation.} Propose using `visit\_webpage' to extract the HTML or text contents of the homepage for archived homepage snapshots of \texttt {orionmindproject.com}.
    \end{tcolorbox}

    \textbf{\faRocket \ Final Execution (Step 16-18):} \\
    $\rightarrow$ \textbf{Action:} \texttt{visit\_webpage} on archived \textbf{source code.} \\
    $\rightarrow$ \textbf{Key Observation:} Detected a hidden textual anchor: \textit{``the alien writing on the top is... Kesovan symbols... see what they mean here.''} \\
    $\rightarrow$ \textbf{Result:} Accessed symbols glossary. Symbol 10 (\textit{Leaning curve}): \textbf{``War is not here this is a land of peace''} \checked
\end{tcolorbox}
\caption{An example of how \method triggers dynamic re-planning, reflects on the plan based on the planning errors, and finally revises the plan to avoid failure.}
\label{fig:qualitative_analysis}
\end{figure*}

\subsection{Main Results}
Table~\ref{tab:main} presents the primary experimental results comparing \method  against various baselines on the GAIA and SimpleQA benchmarks. To ensure a rigorous and equitable comparison, we maintain strictly identical settings, such as tool set and inference parameters, across all evaluated methods, thereby eliminating confounding factors related to environmental configuration.

\noindent\textbf{Quantitative Analysis.} Table~\ref{tab:main} presents the performance of \method compared to representative agentic and reflective baselines on GAIA and SimpleQA benchmarks. On GAIA, \method achieves the highest total score of $58.18\%$ (GPT-4.1) and $59.39\%$ (Gemini-2.5-pro), representing an average improvement of $17.14\%$ and $11.68\%$ over baselines. Besides, the performance gap on Level 3 tasks, where \method outperforms baselines by $14.29\%$ on average, suggests that prospective reflection excels at high-complexity reasoning where error propagation is most severe.

On the SimpleQA benchmark, \method significantly enhances factuality, achieving an average improvement of $12.79\%$ in the Correct metric compared to all baselines across both LLM backbones. The simultaneous increase in the Correct Given Attempted metric ($6.08\%$ on average for GPT-4.1 and $7.73\%$ on average for Gemini-2.5-pro) and decrease in the Not Attempted metric ($12.29\%$ on average for GPT-4.1 and $4.00\%$ on average for Gemini-2.5-pro) indicate that our method allows agents to locate grounded information. Different from traditional methods to improve the factuality of an LLM without tools \citep{wang2025truthflow, cao2024personalized}, the performance gain under the agentic setting demonstrates that our proposed method results in a more comprehensive and robust system that further avoids hallucinations.

\noindent\textbf{Qualitative Analysis.} Figure~\ref{fig:qualitative_analysis} illustrates how \method handles a complex Level-3 GAIA task to \textit{identify the meaning of a specific symbol on Eva Draconis' personal website which can be accessed on her YouTube page}. The task is particularly challenging because both the target YouTube channel and the personal webpage 
is no longer active. While the agent initially struggles for 14 steps with empty pages and missing links, \method triggers a ``turning point'' at step 15 via prospective reflection. Recognizing that standard searches are futile, the agent performs dynamic re-planning to pivot toward the Wayback Machine. Crucially, to mitigate potential \textit{ineffective tool selection} problem, the revised plan shifts from direct image retrieval to source-code excavation. By extracting raw HTML via the ``visit\_webpage'' tool, the agent successfully locates the archived symbol description, completing a task that would otherwise result in a failure state.

\subsection{Comparison with Complex Agent Frameworks}
Beyond the controlled reflection-based comparisons in Table~\ref{tab:main}, Table~\ref{tab:comp_frame} situates \method within a broader landscape of sophisticated agent frameworks. These systems are typically evaluated under non-standardized and substantially more resource-rich settings, often involving more powerful tool access, different backbone models, larger action budgets, and specialized prompt or memory designs.
As a result, their reported performance is not directly comparable to our controlled experimental setting. We therefore cite these results directly from the original publications and present them separately to provide qualitative context rather than a strict head-to-head evaluation. Detailed configurations and sources for these agent frameworks are provided in Appendix~\ref{appendix:source}.

\begin{table}[ht]
\vspace{-5pt}
\caption{Comparative results on the GAIA validation set between various agentic frameworks. Results are directly cited from their respective original reports or other sources.}
\label{tab:comp_frame}
\centering
\resizebox{0.95\linewidth}{!}{
    \begin{tabular}{llcccc}
        \toprule
        \multirow{2}{*}{\textbf{Frameworks}}& \multirow{2}{*}{\textbf{Backbone LLM}} & \multicolumn{4}{c}{\textbf{GAIA (\%)}} \\
        \cmidrule(lr){3-6}
        & & L1 ($\uparrow$) & L2 ($\uparrow$) & L3 ($\uparrow$) & Total ($\uparrow$) \\
        \midrule
        \textbf{AutoAgent} & Claude-Sonnet-3.5 & 71.70 & 53.49 & 26.92 & 55.15 \\
        \textbf{Magnetic-1} & o1 & 56.60 & 46.51 & 23.08 & 46.06 \\
        \textbf{HAL Agent} & GPT-4.1 & 52.83 & 55.81 & 23.08 & 49.70 \\
        \textbf{HF-ODR} & GPT-4.1 & 58.49 & 50.00 & 34.62 & 50.30 \\
        \textbf{HF-ODR} & o1 & 67.92 & 53.49 & 34.62 & 55.15 \\
        \textbf{OWL} & GPT-4.1 & 71.70 & 50.00 & 26.92 & 53.33 \\
        \textbf{OAgent} & GPT-4.1 & 67.92 & 53.49 & 34.62 & 55.15 \\
        \midrule
        \textbf{\makecell[l]{Smolagents \\ + \method}} & GPT-4.1 & 71.70 & 55.81 & 38.46 & 58.18 \\
        \textbf{\makecell[l]{Smolagents \\ + \method}} & Gemini-2.5-pro & 67.92 & 60.47 & 38.46 & 59.39 \\
        \bottomrule
    \end{tabular}
}
\vspace{-10pt}
\end{table}

As shown in Table~\ref{tab:comp_frame}, \method outperforms significantly more complex multi-agent architectures (e.g., AutoAgent and OWL) by over $3\%$. These results underscore that effective prospective reflection can surpass simply increasing execution steps or orchestrating additional agent roles. Importantly, \method is \textbf{orthogonal} to these frameworks and can in principle be integrated as a reflection module to further improve their planning and execution robustness.

\subsection{Transferability}
Although we primarily instantiate \method within the Smolagents framework, we further evaluate its transferability across different agent architectures. Specifically, we integrate \method into a distinct agent framework, OWL~\citep{hu2025owl}. To ensure a fair comparison, we evaluate both variants under the same setting as in our main experiments (identical tool access, execution budgets, and hyperparameter settings), and report pass@1 accuracy on the GAIA validation set.

\begin{table}[ht]
    \caption{Transferability results across different agentic systems.}
    \label{tab:transfer}
    \centering
    \resizebox{0.85\linewidth}{!}{\begin{tabular}{lcccc}
        \toprule
        \textbf{Frameworks} & \textbf{Level 1} & \textbf{Level 2} & \textbf{Level 3} & \textbf{Total} \\
        \midrule
        \textbf{Smolagents} & 56.60 & 47.67 & 19.23 & 46.06 \\
         + \textbf{\method} & 71.70 & 55.81 & 38.46 & 58.18 \\
        \midrule
        \textbf{OWL} & 60.38 & 51.16 & 26.92 & 50.30 \\
         + \textbf{\method} & 73.58 & 58.14 & 42.31 & 60.61 \\
        \bottomrule
    \end{tabular}}
    \vspace{-5pt}
\end{table}

As shown in Table~\ref{tab:transfer}, \method consistently improves performance over the corresponding backbone, indicating that prospective reflection can serve as a general add-on module beyond a single agent implementation.

\subsection{Error Distribution}

\begin{figure}[bh]
    \centering
    \includegraphics[width=0.85\linewidth]{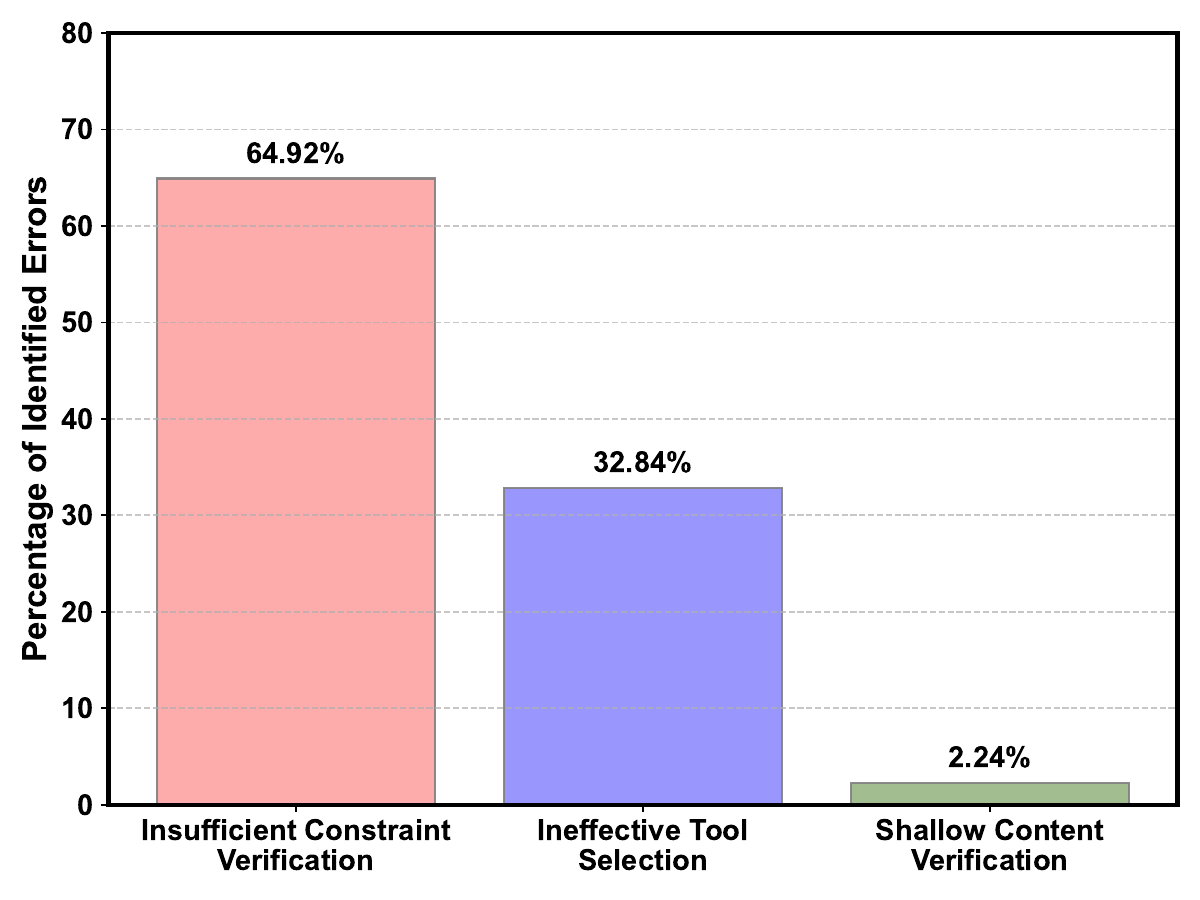}
    \caption{Error distribution of \method on GAIA using GPT-4.1 as the backbone LLM.}
    \label{fig:error_dist}
\end{figure}

To further reveal the error patterns exhibited by our proposed reflection mechanism, we analyze the reflective processes and error recognition of \method with GPT-4.1 backbone on the GAIA validation set. Across all 165 tasks, the defect rate, defined as the proportion of plans identified as risky among all updated plans, is $74.44\%$, revealing that most of the plans are reflected as carrying critical errors. As illustrated in Figure~\ref{fig:error_dist}, the error distribution is heavily skewed: insufficient constraint verification ($64.92\%$) and ineffective tool selection ($32.84\%$) account for nearly all identified errors.

This concentration suggests a fundamental gap in planning. While the agent aims for concise, high-level strategies, the high frequency of constraint-related and tool-related errors reveals that critical task boundaries and reasonable tool selection are often lost in abstraction. This indicates that for complex reasoning tasks, conciseness may come at the cost of operational rigor. Without explicit grounding in task-specific environments, the agent tends to neglect something that can be the key to completing the task correctly.

\subsection{Ablations}

To rigorously assess the individual contributions of the Planning Errors (PE) and Dynamic Re-Planning (DRP) modules, we conducted an ablation study using Smolagents as our baseline framework. The results, summarized in Table~\ref{tab:ablation}, isolate the impact of removing each component while keeping the experimental setting consistent.

\begin{table}[ht]
    \caption{Ablation studies on the planning errors (denoted as PE) and the dynamic re-planning (denoted as DRP). Results are from GPT-4.1 as the backbone LLM on the GAIA validation set.}
    \label{tab:ablation}
    \centering
    \resizebox{0.9\linewidth}{!}{\begin{tabular}{lcccc}
        \toprule
        \textbf{Method} & \textbf{Level 1} & \textbf{Level 2} & \textbf{Level 3} & \textbf{Total} \\
        \midrule
        Smolagents & 56.60 & 47.67 & 19.23 & 46.06 \\
        \method w/o PE & 58.49 & 53.49 & 23.08 & 50.30 \\
        \method w/o DRP & 54.72 & 54.65 & 34.62 & 51.52 \\
        \method & 71.70 & 55.81 & 38.46 & 58.18 \\
        \bottomrule
    \end{tabular}}
\end{table}

\noindent\textbf{Planning Errors as Empirical Priors.} The ablated results indicate that the Planning Errors is essential for the transition from retrospective to prospective reflection. Removing it leads to a significant performance drop, particularly on Level 3 tasks (decreasing from $38.46\%$ to $23.08\%$). This aligns with our hypothesis that predicting future errors is inherently difficult without historical context. PE acts as an ``empirical prior'', grounding the prospective reflection and allowing the agent to anticipate and avoid potential pitfalls before execution begins.

\noindent\textbf{Dynamic Re-Planning as Compensation.} While PE aids in pre-execution error avoidance, the \textit{w/o DRP} variant reveals the necessity of runtime flexibility. Despite having access to historical error experiences, the agent may still fail to detect or mitigate every risk during the planning phase. The $6.66\%$ total performance gap between \method and \textit{w/o DRP} demonstrates that re-planning serves as a critical safety net, without which, the added complexity of prospective reflection might lead to rigid plans that the agent cannot recover from if a minor execution error occurs.

\subsection{Cost Analysis}
\begin{figure}[ht]
    \centering
    \includegraphics[width=0.95\linewidth]{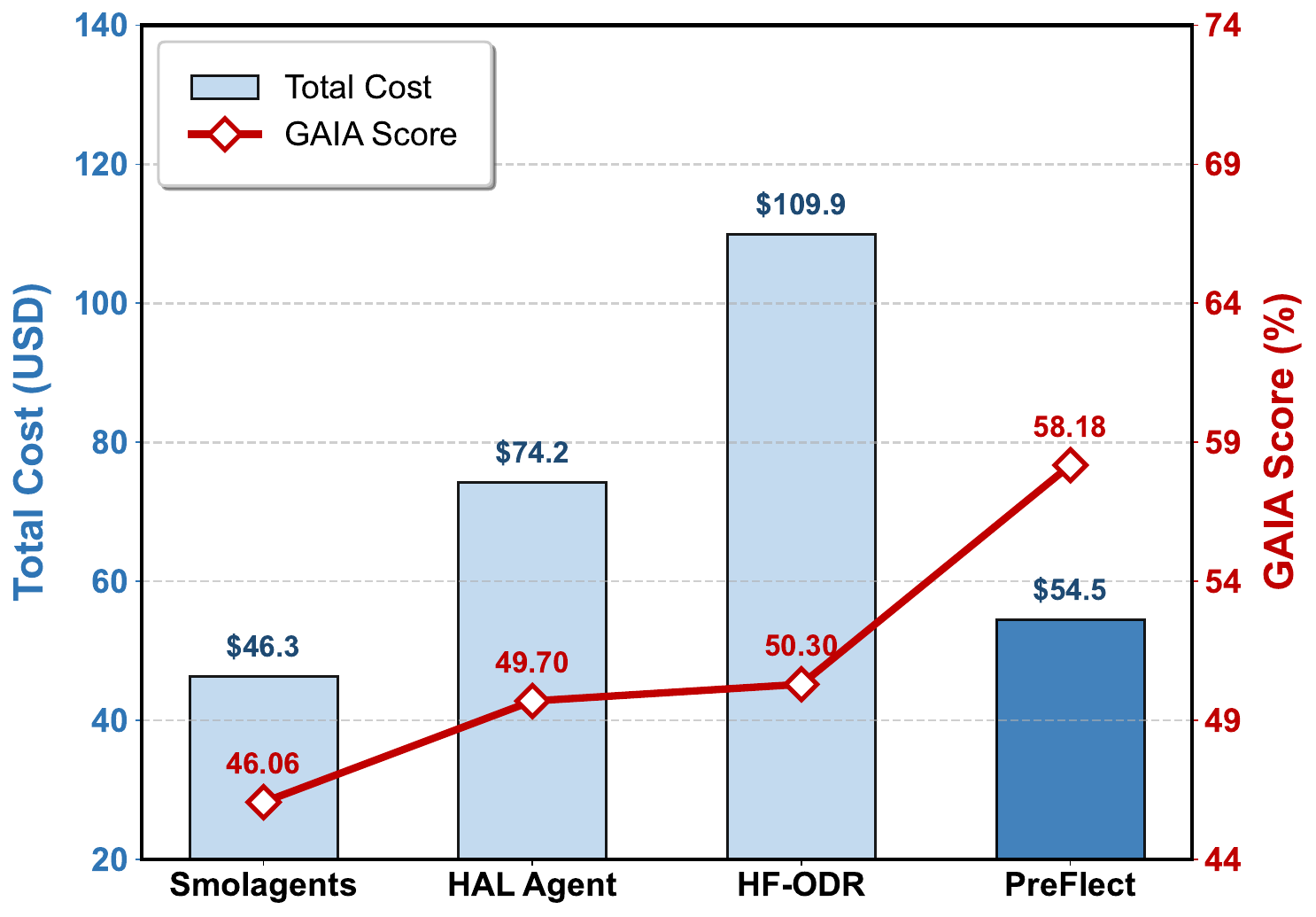}
    \caption{Performance-cost trade-off on the GAIA validation set. The primary axis shows the total cost (USD) and the secondary axis shows the corresponding GAIA scores.}
    \vspace{-10pt}
    \label{fig:cost}
\end{figure}

To evaluate the practical scalability of \method, we analyze its operational costs relative to performance on the GAIA validation set. We compare results of GPT-4.1 as the backbone LLM whose input price is \$2 per million tokens while the output price is \$8 per million tokens. As illustrated in Figure~\ref{fig:cost}, \method achieves a superior performance-cost trade-off compared to existing frameworks. While the baseline Smolagents costs \$46.31, our method reaches a highest score of 58.18\% with only a marginal 17.64\% cost increase (\$54.48). In stark contrast, advanced agents like HAL and HF-ODR incur significantly higher expenditures (\$74.19 and \$109.88, respectively) yet yield lower performance (49.70\% and 50.30\%).

\section{Conclusion}
\label{sec:conclusion}

In this work, we propose \method, a prospective reflection mechanism that shifts from reactive correction to proactive failure prevention. Our approach introduces distilled planning errors to provide structured and grounded error detection, while a dynamic re-planning mechanism brings up an updated plan in case errors are neglected during execution. Experimental results demonstrate that \method's significant improvements in complex task solving and remarkable transferability across different agent architectures.








\newpage
\section*{Impact Statement}

This work introduces a novel prospective reflection mechanism to enhance general capability of LLM agents. By addressing significant weaknesses of previous reflection methods, \method contributes to the development of more reliable and advanced agent systems.


\bibliography{example_paper}
\bibliographystyle{icml2026}

\newpage
\appendix
\onecolumn

\section{Implementation Details}
\label{appendix:imp}

\subsection{Agent Framework}
We build \method based upon Smolagents \citep{smolagents} framework. Specifically, we add the prospective reflection mechanism and dynamic re-planning to the \texttt{CodeAgent} class. Therefore, the agent that executes all the tool calls and reasonings is a single agent. Due to resource limitations, we set the maximum action steps to 20. For re-planning, we enforce at least 2 extra planning steps to ensure there's room for the agent to perform periodic summarization and planning. Therefore, except for the initial planning stage, if there is not at least one updated planning step between the 1st to 6th steps, and 7th to 12th steps, we enforce the agent to update the plan at the 6th and 12th steps. The prompt for planning follows the default prompt in Smolagents for the \texttt{CodeAgent} class.

\subsection{Tools}
We use the minimum necessary tools for both GAIA and SimpleQA, and all the experiments except Table~\ref{tab:comp_frame} are done with the same tool set.
\begin{itemize}
    \item Web Search: Due to resource limitations, we use the default web search tool in Smolagents and use the DuckDuckGoSearch as the search engine.
    \item Visit Webpage: Similarly, we use the default visit webpage tool in Smolagents, which is powered by `requests' and `markdownify' libraries.
    \item Text Inspector: This tool enables the agent to parse diverse document types (e.g., .pdf, .docx, .pptx, .xlsx, .html) into Markdown text via a MarkdownConverter. It supports both raw content retrieval and query-based analysis, returning structured responses with brief answers, detailed analysis, and document context. This tool can also be found in Smolagent's example of using the ODR agent to run GAIA tasks.
    \item Visual Inspector: Designed for image processing (.jpg, .png, .gif, .bmp), this tool utilizes a Vision-Language Model to interpret visual data. It automatically generates detailed captions or performs targeted visual question-answering by encoding images into Base64 for API-based inference.
    \item Audio Inspector: This tool facilitates auditory analysis by transcribing files (.mp3, .m4a, .wav) using the OpenAI Whisper-1 model \citep{radford2023robust}. It provides either a direct transcription or a structured three-part response (brief, detailed, and contextual) to specific natural language questions regarding the audio content.
\end{itemize}

For the LLM-powered tools, we uniformly adopt Gemini-3-flash-preview (i.e., \texttt{gemini-3-flash-preview}) for all the experiments in order to ensure the most advanced multi-modal processing ability.

\subsection{Reflection}

We utilize a separate agent (i.e., the reflector) to handle prospective reflection in order to achieve maximum lightweight and clear context. Specifically, whenever a plan is generated, we call an agent powered by the same backbone LLM to execute a multi-turn conversation where both plan reflection and revision are completed. The system prompt specifies the planning errors and instructions, and assigns the role of a reflector to the agent. In the first turn of conversation, we input the task, plan, available tools, and a summarization of past actions, and expect to obtain an analytical diagnosis of the plan. If any critical error is detected, we enter the second turn to revise the plan, utilizing the success and failure patterns in the planning errors as a reference. All the prompts are provided in Appendix~\ref{appendix:prompt}.

\subsection{LLM Settings}
We call OpenAI API to use GPT-4.1 model (i.e., \texttt{gpt-4.1-2025-04-14}) and use the \texttt {OpenAIModel} class in Smolagents to load the model, with exactly the default parameters across all the experiments. Similarly, we utilize the \texttt{LiteLLMModel} class to load Gemini models, such as Gemini-2.5-pro (i.e., \texttt{gemini-2.5-pro}), and also follow the default parameters.

As for the SimpleQA evaluation pipeline, which adopts an LLM-as-a-judge fashion, we utilize GPT-4o (i.e., \texttt{gpt-4o-2024-08-06}) to determine whether the agent's answer is correct, incorrect, or not attempted.



\section{Planning Errors}
\label{appendix:error}

All the LLMs we use to construct the planning errors are Gemini-3-pro-preview (i.e., \texttt{gemini-3-pro-preview}) due to its strong long-context understanding and processing ability.

\subsection{Construction}
We randomly select 50 hard-level data samples from the HotpotQA \citep{yang2018hotpotqa} training set and 50 samples from the MuSiQue \citep{trivedi2022musique} training set as our seed tasks. By repeatedly running the base agent (i.e., Smolagents with GPT-4.1) on these tasks to collect three trajectories per task, we obtain 300 trajectories in total. However, we only particularly retain tasks involving both successful and failure trajectories as mentioned in Section~\ref{subsubsec:plan_error}. This results in a final candidate subset of 32 tasks, including 18 from HotpotQA and 14 from MuSiQue.

\begin{tcolorbox}[
    colback=gray!5,
    colframe=gray!50!black,
    title=\textbf{Example Diagnosis from HotpotQA},
    fonttitle=\bfseries,
    breakable
]

\textbf{Task:} What character did this actor, who also played a supporting role in Sleepy Hollow, play in the final six Harry Potter films?

\tcbline

\textbf{Diagnosis:}
\vspace{-8pt}

\begin{description}[
    leftmargin=0.0em, 
    font=\bfseries\itshape, 
    itemsep=-0.5em        
]
    \resultitem{Error Type:}{\texttt{\textcolor{red!70!black}{inadequate cross verification of constraints}}}
    
    \resultitem{Description:}{The agent identifies a candidate answer but fails to rigorously verify if it meets all the specific constraints defined in the task (e.g., `final six films').}
    
    \resultitem{Evidence:}{TRAJ 2, Step 19: ... ``There is no evidence ... of any other Sleepy Hollow supporting actor being present in the last six Harry Potter films. The consistent answer is Richard Griffiths as Vernon Dursley.''}
    
    \resultitem{Impact:}{In Trajectory 2, the agent identifies Richard Griffiths but fails to verify if he is in \textit{all} of the `final six' films. In contrast, Trajectory 1 and Trajectory 3 correctly identify Michael Gambon, who ..., fitting the constraint.}

\end{description}

\end{tcolorbox}

\begin{tcolorbox}[
    colback=gray!5,
    colframe=gray!50!black,
    title=\textbf{Example Diagnosis from MuSiQue},
    fonttitle=\bfseries,
    breakable
]

\textbf{Task:} When did the torch arrive in the region where Irreversi takes place?

\tcbline

\textbf{Diagnosis:}
\vspace{-8pt}

\begin{description}[
    leftmargin=0.0em,      
    font=\bfseries\itshape, 
    itemsep=-0.5em    
]

    \resultitem{Error Type:}{\texttt{\textcolor{red!70!black}{tool selection inflexibility}}}
    
    \resultitem{Description:}{The agent continues to use a specific tool (e.g., `visit\_webpage') on a specific domain (e.g., Wikipedia) despite repeated failures (403 Forbidden), instead of switching to alternative tools or alternative sources immediately.}
    
    \resultitem{Evidence:}{TRAJ 2, Step 13: The agent attempts to use `visit\_webpage' on the Wikipedia URL ... despite knowing from Step 4 that Wikipedia blocks this tool.}
    
    \resultitem{Impact:}{This error prevents TRAJ 2 from accessing the definitive source of the schedule. TRAJ 1 encounters the same 403 error initially but later correctly switches to `inspect\_file\_as\_text' (TRAJ 1, Step 22) to read the Wikipedia page and successfully retrieves the date. TRAJ 3 avoids Wikipedia entirely for the date confirmation and successfully uses a news site (CNN).}

\end{description}

\end{tcolorbox}

\subsection{Final Planning Errors} 

We prompt an LLM to generate diagnoses by comparing and analyzing the trajectories. The 32 tasks eventually lead to 68 diagnoses in total, with various identified error types and detailed analyses. However, most of these errors have similar essences and therefore can be merged together. As mentioned in Section~\ref{subsubsec:plan_error}, we also perform a manual filter to drop the errors that are not related to long-term planning or too specific to tasks. Specifically, after LLM aggregation, the planning errors contain four error types: insufficient constraint verification, shallow content verification, ineffective tool selection, and ineffective search query. Among these error types, the ``ineffective search query'' is too specific to action-level reasoning instead of high-level planning. During the planning stage, the agent tends to propose a general guideline rather than every detailed action to take. Therefore, we remove the ineffective search query error from the planning errors and end up with three error types covering aspects of constraint, tool, and content understanding.

Particularly, the planning error construction is an \textbf{offline} process which is separated from the major agent reflection and working process. We sample various trajectories from basic Smolagents with GPT-4.1 as the backbone to build up planning errors and directly reuse them with \method powered by either GPT-4.1 or Gemini-2.5-pro. The performance boosts across different backbone LLMs reflect the generalizability of the experiential prior, further validating the effectiveness of the grounded guidance in detecting potential risks in planning stage.

For error identification, we use the error types, descriptions, and 2 failure examples as the experiential prior to help the reflector perform a more structured and precise diagnosis. As for plan revision, we provide the reflector with all the success and failure examples corresponding to the identified error so that the reflector accesses more high-quality references extracted from the agent's own experience.

\newtcolorbox{TaxonomyBox}[1]{
    colback=white, 
    colframe=gray!60, 
    fonttitle=\bfseries,
    colbacktitle=gray!15,
    coltitle=black,
    enhanced,
    attach boxed title to top left={yshift=-3mm, xshift=3mm},
    boxed title style={colframe=gray!60, sharp corners},
    title=#1,
    arc=0mm,
    breakable,
    before skip=15pt,
    after skip=15pt
}

\newcommand{\failureexample}[2]{
    \begin{quote}
        \small\color{gray!80!black}
        \textbf{Failure #1:} #2
    \end{quote}
}

\begin{TaxonomyBox}{Planning Error Details}
\begin{description}[style=nextline, leftmargin=1em, font=\bfseries\ttfamily\color{blue!70!black}]

    \item[insufficient\_constraint\_verification] 
    The plan identifies a potential answer but fails to rigorously verify it against all specific constraints or conditions in the prompt (e.g., timeframes, specific subsets, exclusions). It relies on partial matches, ambiguous snippets, or initial findings without including steps to explicitly confirm that the candidate satisfies every requirement using authoritative sources before concluding.
    
    \failureexample{Example 1}{In a task identifying an actor who appeared in `Sleepy Hollow' and the `final six' Harry Potter films, the agent identified Richard Griffiths (Vernon Dursley) because he appeared in both franchises. However, the plan failed to verify his presence in *each* of the final six films (he is absent from some), leading to an incorrect answer based on a partial match.}
    \failureexample{Example 2}{In a task identifying the 2004 sporting event participated in by the star of `Uyirodu Uyiraga' (Ajith Kumar), the agent relied on a single ambiguous news snippet mentioning `Formula Asia BMW F3 Championships' and ignored a previous snippet mentioning `British Formula 3'. It failed to visit a dedicated database to resolve the conflict or verify the specific year, leading to an incorrect answer.}

    \item[shallow\_content\_verification] 
    The agent relies solely on search snippets or broad summaries without visiting the actual webpages, or extracts data from visited pages without verifying if the specific data definitions (e.g., geographic scope, units) match the task constraints. It fails to recognize that generic pages often serve as navigation hubs or that specific labels in the text might contradict its assumptions.
    
    \failureexample{Example 1}{The agent repeatedly searched for `2008 Olympic torch relay Hong Kong date' but dismissed the search snippets as `general year summaries' without visiting the pages to find the specific date within the text}
    \failureexample{Example 2}{The agent relied too heavily on the \texttt{web\_search} text output or a blocked Wikipedia page, missing the opportunity to extract the answer from other accessible sources (like a Daily Mail article) visible in the search results.}

    \item[ineffective\_tool\_selection] 
    The agent persists in using a specific tool or approach despite known limitations or repeated failures (e.g., using a web scraper on blocked sites, or repeatedly searching for a direct answer when results are noisy), failing to switch to alternative tools or sources (e.g., visiting a homepage to navigate manually).
    
    \failureexample{Example 1}{The agent attempted to use \texttt{visit\_webpage} on a Wikipedia URL to find the torch relay schedule, despite receiving a 403 Forbidden error, and did not attempt to use \texttt{inspect\_file\_as\_text} which can handle such URLs.}
    \failureexample{Example 2}{The agent repeatedly attempted to visit a Wikipedia page despite receiving 403 Forbidden errors, and even tried to inspect a non-existent local path derived from the URL, rather than pivoting to other accessible news articles found in search results.}

\end{description}
\end{TaxonomyBox}

\section{Agent Sources}
\label{appendix:source}

The results of both the HAL Generalist Agent and the HF-ODR are directly cited from HAL's website \footnote{\url{https://hal.cs.princeton.edu/gaia}}, including GAIA scores and total costs. OAgent's result is from the original paper \citep{zhu2025oagents} while OWL's result is reproduced by \citet{wang2025efficient}. We also directly cite the results of AutoAgent and Magnetic-1 from \citet{chen2023autoagents}.

\subsection{Details of the Agent Frameworks}
In this section, we expand on the implementation details of other open-source complex agent frameworks. By analyzing the engineering details, we further claim that our proposed method achieves higher performance even with limited computing resources, validating the effectiveness of the proposed reflection mechanism.

\begin{itemize}
    \item \textbf{HAL Generalist Agent:} It is a single-agent framework that is built upon Smolagents' CodeAgent class with 200 max action steps and a fixed interval planning update of every 4 steps. The agent has access to the GoogleSearchTool, the default VisitWebpageTool and PythonInterpreterTool, TextInspectorTool, execute bash, edit file, file content search, and a visual inspector tool powered by GPT-4o.
    \item \textbf{HF-ODR:} It is a multi-agent framework that consists of a manager agent and a text browser agent, which is specialized in browsing webpages. The browser agent is allowed to execute up to 20 action steps with a fixed interval of planning at every 4 steps. It is also equipped with comprehensive tools to browse webpages, including GoogleSearchTool, VisitTool, page scrolling (PageUpTool and PageDownTool), FinderTool (i.e., Ctrl+F), FindNextTool (i.e., finding the next match in a Ctrl+F search), ArchiveSearchTool (i.e., search the Wayback Machine), and TextInsepctorTool. The manager agent, equipped with a text inspector tool and a visual inspector tool, has 12 maximum action steps and a plan every 4 steps.
    \item \textbf{OAgents:} It has a similar architecture to HAL Generalist Agent, where a single CodeAgent handles everything. Differently, it employs the optimal modules that are empirically validated in their experiments. The system supports up to 100 maximum action steps. Its core architecture features a multi-faceted planning module that supports subtask decomposition, staged plan calibration, and periodic plan reviews every 1 step. For factual grounding, it utilizes an ensembled toolkit comprising a multi-source search engine (Google, Wikipedia, Baidu, Bing, DuckDuckGo) alongside specialized multimodal inspectors for text, audio (Whisper-1), and visual analysis. The framework further incorporates a hierarchical memory system (summarization, vectorized retrieval, and long-term memory fusion) and a Test-Time Scaling (TTS) module that employs Best-of-N sampling and list-wise verification to enhance decision-making robustness.
    \item \textbf{Magentic-One:} It features an orchestrator agent that implements an outer loop to manage the task ledger and an inner loop to manage the progress ledger. Except for the orchestrator, the system consists of a \textit{WebSurfer} to manage web browser, a \textit{FileSurfer} to read local files of most types, a \textit{Coder} to write code, and a \textit{ComputerTerminal} to provide access to a console shell where the Coder’s programs can be executed, and where new programming libraries can be installed.
    \item \textbf{AutoAgent:} This framework utilizes a fully automated multi-agent architecture where a central LLM-powered actionable engine orchestrates specialized worker agents, including Web, Code, and File agents. It employs a Manager-Worker inspired by Magnetic-One, supporting dynamic task decomposition and sequential reasoning. The self-play agent customization module enables autonomous generation and refinement of tools, agent profiles, and workflows with natural language. The framework also integrates an agentic RAG system to ingest diverse file formats such as PDF and images.
    \item \textbf{OWL:} OWL is a multi-agent system built upon Camel with a Role-Playing architecture. Two agents, respectively, play the roles of user and assistant to solve user input tasks. The assistant agent manages various tools and specialized agents, including a \textit{Browser} system that is composed of a browser agent and a planning agent, \textit{multi-modality handling} tools including image, video, and audio, \textit{CodeExecution} tool, \textit{Excel} tool, \textit{Search} tool, and \textit{FileWrite} tool. The user agent decomposes the task and prompts the assistant agent to finish a sub-task while the assistant agent assigns specialized agents to execute and provides execution results or failures to the user agent.
\end{itemize}

\section{Prompts}
\label{appendix:prompt}
\definecolor{promptgray}{RGB}{245, 245, 245}
\definecolor{bordergray}{RGB}{200, 200, 200}
\definecolor{titleblue}{RGB}{40, 60, 110}

\definecolor{titlegreen}{RGB}{46, 139, 87}
\definecolor{variablecolor}{RGB}{180, 50, 50}

\lstset{
    basicstyle=\ttfamily\small,
    breaklines=true,
    breakatwhitespace=true,
    columns=flexible,
    keepspaces=true,
    extendedchars=false,
    inputencoding=utf8
}

\newtcolorbox{PromptBox}[1]{
    colback=promptgray,       
    colframe=bordergray,      
    fonttitle=\bfseries\color{white},
    coltitle=white,
    colbacktitle=titleblue,   
    enhanced,
    attach boxed title to top left={yshift=-2mm, xshift=2mm},
    boxed title style={sharp corners, size=small},
    title=#1,                 
    arc=1mm,                  
    boxrule=0.5pt,            
    breakable,                
    fontupper=\small\ttfamily, 
    before skip=10pt,         
    after skip=10pt           
}

\definecolor{instpurple}{RGB}{100, 50, 150} 
\definecolor{exgray}{RGB}{240, 242, 245}  
\definecolor{codeblue}{RGB}{30, 80, 160} 
\definecolor{userblue}{RGB}{40, 90, 150}

\newtcolorbox{SysPromptBox}[1]{
    colback=instpurple!5,
    colframe=instpurple,
    fonttitle=\bfseries,
    colbacktitle=instpurple,
    title=#1,
    enhanced,
    attach boxed title to top left={yshift=-2mm, xshift=2mm},
    boxed title style={sharp corners},
    arc=1mm,
    breakable
}

\newtcolorbox{UserPromptBox}[1]{
    colback=userblue!5,           
    colframe=userblue,           
    fonttitle=\bfseries,
    colbacktitle=userblue,       
    title=#1,
    enhanced,
    attach boxed title to top left={yshift=-2mm, xshift=2mm},
    boxed title style={sharp corners},
    arc=1mm,
    breakable
}

\newtcolorbox{ExampleBox}[1]{
    colback=exgray!50,
    colframe=gray!50,
    fonttitle=\bfseries\color{black},
    colbacktitle=gray!20,
    title=#1,
    enhanced,
    attach boxed title to top left={yshift=-2mm, xshift=2mm},
    boxed title style={sharp corners, colframe=gray!50},
    arc=0mm,
    boxrule=0.5pt,
    breakable,
    fontupper=\small
}

\newcommand{\promptcode}[1]{
    \begin{tcolorbox}[
        colback=gray!10, 
        colframe=gray!30, 
        boxrule=0.2pt, 
        left=2pt, right=2pt, top=2pt, bottom=2pt,
        fontupper=\small\ttfamily\color{codeblue}
    ]
    #1
    \end{tcolorbox}
}

\newcommand{\promptvar}[1]{\textcolor{variablecolor}{\texttt{\{#1\}}}}


\subsection{Prompt For Prospective Reflection}

\begin{SysPromptBox}{System Prompt: Reflection \& Revision}
You are a hyper-vigilant AI Quality Assurance agent designed to perform pre-execution strategic reflection on an agent's plan to prevent task failure and wasted effort. Your analysis should be bi-directional: look backward to review the past actions and forward to prevent future errors.

\smallskip
Your role in this multi-turn conversation:
\begin{enumerate}[leftmargin=*, nosep]
    \item First turn: Detect critical errors in the updated plan based on the provided error taxonomy
    \item Second turn: Revise the plan to address the detected errors
\end{enumerate}

\smallskip
You must focus only on the \textbf{critical errors} that fundamentally lead to task failure and ignore minor errors that are hardly harmful or easily fixable with minor adjustments.

\smallskip
\#\# Error Taxonomy \\
\promptvar{error\_taxonomy}

\smallskip
\#\# Core Principles
\begin{itemize}[leftmargin=*, nosep]
    \item \textbf{Pragmatism over Pedantry}: Only flag at most two most critical errors that will *actually* cause failure.
    \item \textbf{Maximize Task Success}: Focus on flaws that have a real risk of causing failure or incorrect results.
    \item \textbf{Root Cause Focused}: Identify the underlying cause, not just surface manifestations.
    \item \textbf{Trajectory Awareness}: Use the history summary to diagnose recurring or systemic failures.
    \item \textbf{Reference}: Use failure/success examples to understand error types and solutions.
    \item \textbf{Feasibility and Completeness}: Ensure the plan is executable given available tools.
    \item \textbf{Constraint Awareness}: Pay attention to past mistakes that reveal tool/environment limitations.
\end{itemize}
\end{SysPromptBox}

\begin{UserPromptBox}{User Prompt: Prospective Reflection (Turn 1)}
\# Task\\
\promptvar{task}

\smallskip
\# Trajectory\\
\promptvar{previous\_context}

\smallskip
\# Available Tools\\
\promptvar{available\_tools}

\smallskip
\# Current Plan\\
\promptvar{plan\_content}

\smallskip
\# Instructions for Turn 1: Error Detection\\
Analyze the above plan for critical errors that could lead to task failure.

\smallskip
Your responsibilities:
\begin{enumerate}[leftmargin=*, nosep]
    \item \textbf{Check the plan using the Error Taxonomy}.
    \item \textbf{Enforce learning from history}: Ensure the plan incorporates lessons learned from the trajectory.
\end{enumerate}

\smallskip
Be strict but pragmatic: flag only issues that materially threaten success. If there are multiple critical errors, flag the most critical one first.

\smallskip
However, if the trajectory shows that the agent has already tried everything within capabilities, then you should be lenient with the 'flawed' plan, such as resorting to secondary sources and broadening the scope of search which may contradict the constraints.

\smallskip
\# Output Format\\
Your response MUST be a single JSON object. No other text or explanation is allowed.

\smallskip
\begin{itemize}[label={}, leftmargin=1em]
\item \{
\item \quad "analysis\_result": "PASS" | "FAIL",  // Use "FAIL" ONLY if at least one error from the taxonomy is present.
\item \quad "detected\_errors": [
\item \quad \quad // If "analysis\_result" is "PASS", this array MUST be empty.
\item \quad \quad \{
\item \quad \quad \quad "error\_type": "specific\_error\_type\_from\_taxonomy",
\item \quad \quad \quad "description": "A concise, clear explanation...",
\item \quad \quad \quad "evidence": "A direct quote...",
\item \quad \quad \quad "how\_error\_affects\_task\_completion": "A specific explanation..."
\item \quad \quad \}
\item \quad ]
\item \}
\end{itemize}
\end{UserPromptBox}

\begin{UserPromptBox}{User Prompt: Plan Revision (Turn 2)}
\# Turn 2: Plan Revision

Based on the errors you detected, now revise the plan to address them. The detected errors are in the order of severity. Thus you should revise the plan to prevent the most critical one first. 

\smallskip
\# Evidence\\
\promptvar{error\_evidences\_str}

\smallskip
\# Core Directives for Plan Revision:
\begin{enumerate}[leftmargin=*, label=\textbf{\arabic*.}, itemsep=2pt, parsep=0pt]
    \item \textbf{Prioritize Feedback:} Your absolute top priority is to meticulously analyze and address the most critical errors you detected.
    \item \textbf{Perform Root Cause Analysis:} Do not just fix the symptoms. Before creating a new plan, you must first understand \textit{why} the previous plan and past trajectory failed.
    \item \textbf{Balance Incremental and Substantial Rework:} Take a balance between incremental rework and drastic strategy changes. No need to start from scratch if the original plan is already good.
    \item \textbf{Ensure Feasibility \& Completeness:} The new plan must be robust and executable. \textit{Note toolkit limitations:} \promptvar{available\_tools}.
    \item \textbf{Draw Experience from Previous Errors:} Avoid making similar mistakes. For example, if a Wikipedia page was inaccessible, do not try the same domain again. Explicitly emphasize this in your revised plan.
    \item \textbf{Follow The Plan Structure:} Your improved plan should follow the original structure: \texttt{Facts Survey} and \texttt{Plan} sections.
\end{enumerate}

\smallskip
\# Instructions\\
Generate a revised, improved plan that addresses critical errors. Your output should be the \textbf{complete revised plan text only}, without any JSON formatting or additional explanation.
\end{UserPromptBox}

\subsection{Prompt For Agentic Re-Planning}
In order to leverage the autonomy of the agent, we add an instruction and two examples of dynamic planning into the system prompt.

\begin{SysPromptBox}{System Guidelines: Dynamic Planning}
At each step, in the \texttt{Thought:} sequence, you should first take a holistic view of the history, thinking of whether the current path is optimal. If not, consider calling \texttt{re\_plan()} function to pivot to a new plan. Otherwise, explain your reasoning towards solving the task and the tools that you want to use.

\smallskip
\noindent\textbf{CRITICAL STRATEGIC PRINCIPLE - RE-PLANNING:} 
Since you plan forward, you need to follow your plan. However, the real world is unpredictable. \textbf{Do NOT blindly follow the plan.} Use \texttt{re\_plan()} to make a strategic pivot if the current plan is not working.

\smallskip
\noindent\textbf{HOW TO USE RE-PLANNING:}
If you want to request a re-planning step, you must call \texttt{re\_plan()} in the code sequence. 
\promptcode{
replan\_result = re\_plan("Subtask 'fetch wikipedia page' failed due to 403 Forbidden error... Therefore, I need to re-plan.") \\
print(replan\_result)
}

\textbf{IMPORTANT}: Never propose a new strategy inside \texttt{re\_plan()}. You MUST only state the difficulties encountered!
\end{SysPromptBox}

\begin{ExampleBox}{Examples: Strategic Re-planning}
\textbf{Task:} ``Find the population of the capital city of the country where Mount Everest is located.''

\smallskip
\textit{Thought:} According to my plan, I should first find which country Mount Everest is in...
\promptcode{print(web\_search(query="Mount Everest location country"))}

\textit{Observation:} Mount Everest is located on the border between Nepal and Tibet (China).

\smallskip
\textit{Thought:} The observation breaks my original plan's assumption... it is shared by two countries. The original plan is now ambiguous. I must re-plan to handle this multi-country scenario.
\promptcode{
replan\_result = re\_plan("Critical Ambiguity: Mount Everest is on the border of TWO countries, invalidating the single-country assumption.") \\
print(replan\_result)
}

\vfill
\hrule
\vspace{10pt}

\textbf{Task:} ``Download the latest earnings report from Tesla's investor relations page...''

\smallskip
\textit{Thought:} I will start by finding the Tesla investor relations page URL...
\promptcode{print(web\_search(query="Tesla investor relations earnings reports"))}

\textit{Observation:} Found Tesla investor relations page: \url{https://ir.tesla.com/}

\smallskip
\textit{Thought:} Now I will attempt to visit the page...
\promptcode{print(visit\_webpage(url="https://ir.tesla.com/"))}

\textit{Observation:} Error: 403 Forbidden - Access denied.

\smallskip
\textit{Thought:} I have hit a hard technical blocker... Continuing with the original plan is futile. I need to pivot my strategy to obtain the data from a secondary source.
\promptcode{
replan\_result = re\_plan("Tool Failure: Cannot access primary source (Tesla IR page returns 403 Forbidden).") \\
print(replan\_result)
}
\end{ExampleBox}

\subsection{Prompt For Diagnosis Construction}

\begin{SysPromptBox}{System Prompt: Diagnosis}
\begin{lstlisting}
You are a professional error analyst. You are given three trajectories from an AI agent to solve a same task. Your goal is to compare the three trajectories in a **global perspective** and provide a detailed analysis of the **critical** errors that could have been solved in the **planning phase**. Your diagnosis will be used to build an error taxonomy which contains 1) common critical error types that significantly lead to task failures 2) contrastive success-failure cases that are caused by different error types.
You can ignore some minor errors and only focus on the critical errors that lead to the failure of the task. Besides, you should be tolerant of the early-stage errors since the agent is still exploring the task. If these early errors are fixed later, they should not be diagnosed as critical errors.

## Get To Know the Agent

### Agent Workflow
The agent will first write a plan and update it at every 5 actions, and then execute the plan in action steps which contains 'Thought', 'Action', and 'Observations'. It takes the observations from the environment as input to decide the next action step.
NOTE: The "Thought" in each ACTION step is not a plan step!

### Available Tools
- web_search: Performs a duckduckgo web search based on input query; then returns the top search results. This tool is powered by `ddgs`. It uses `DDGS().text` method to search the web.
- visit_webpage: Visits a webpage at the given url and reads its content as a markdown string. NOTE: This tool is powered by `markdownify` and `requests`, which means it can only handle text-based webpages. For json-heavy webpages, it always returns an error. For example, it returns 403 Forbidden Error for all wikipedia pages.
- inspect_file_as_text: This tool is powered by gpt-4o to read a file as markdown text and ask questions about it. This tool handles the following file extensions: ['.html', '.htm', '.xlsx', '.pptx', '.wav', '.mp3', '.m4a', '.flac', '.pdf', '.docx'], and all other types of text files.
- inspect_file_as_image: This tool is powered by gpt-4o to process image files and answer related questions. This tool supports the following image formats: ['.jpg', '.jpeg', '.png', '.gif', '.bmp'].
- inspect_file_as_audio: This tool uses OpenAI whisper-1 to get audio transcriptions and uses gpt-4o to answer related questions. This tool supports the following audio formats: ['.mp3', '.m4a', '.wav'].

**CRITICAL NOTE ABOUT INSPECTOR TOOLS**: The current environment does NOT support downloading files or accessing local file paths. All inspector tools (inspect_file_as_text, inspect_file_as_image, inspect_file_as_audio) MUST be called with internet-accessible URLs, NOT local file paths. Extremely cautious about using paths like '/path/to/file.png' or './local/file.pdf' - these will fail unless you are completely sure the file is stored locally. Instead, provide URLs that are publicly accessible on the web (e.g., 'https://example.com/file.pdf'). 

## Diagnosis

### Purpose of the Diagnosis
The diagnosis will be aggregated into an error taxonomy for agents to reflect on their plans. For future tasks, the agent will self-check its plans and action steps so far, use this error taxonomy to identify critical errors in the plan, and revise the plan to guide the agent to the optimal path given task and environment constraints. Thus you should do everything to serve this purpose.

### Diagnosis Instructions
1. Compare the three trajectories and understand the different and similar details between them.
2. Identify the key differences between them that lead to the different outcomes.
  - The errors should be **GENERAL** enough to be applied to other tasks
  - Make sure that the error can be identified and fixed right after the flawed plan is generated
3. Given the current available tools, make sure that the success is achievable by fixing the critical errors instead of by luck.
4. Extract error patterns that are generalizable to other tasks. Make sure that for new tasks, these error types can be applied to diagnose the critical errors and guarantee the success of new tasks.
5. Focus on three significant parts of the trajectory: the plan, the thought, and the tool usage. Your error types should correspond to them. For example, a 'logical_flawed_plan' error can refer to a plan that has significant flaws in logics; an 'overlooked_observation' error may infer that in the 'Thought', the agent ignores some important observations; a 'tool_failure_ignorance' error may happen when the agent fails to rectify the tool usage in an 'Action'.
**EXAMPLE:** For three trajectories with the same task to find the rating of the first featured review on IMDb page for 'Iron Man', TRAJ 1 successfully gets the page url 'https://www.imdb.com/title/tt0371746/', and then uses the `inspect_file_as_text` tool to read the webpage content and get the rating. However, TRAJ 2 and TRAJ 3 get stuck in visiting the correct url with `visit_webpage` tool. A 403 Forbidden Error occurs repeatedly while the agent keeps using `visit_webpage`. In this case, the error type can be 'tool_failure_ignorance' since the agent fails to identify the 403 error with `visit_webpage` tool and does not plan to alternate to use `inspect_file_as_text`.

### Diagnosis Format
You should organize your analysis into a JSON format with the following fields:
- **error_type**: the type of critical error made by the agent in those failure trajectories. It should be very **GENERAL** with no specific details or information about this task. For example, if the agent ignores the constraint in the task which limits the search range to be between 2020 and 2023, the error type can be 'constraint_ignorance'. If the agent fails to visit the webpage with `visit_webpage` tool and does not realize that `inspect_file_as_text` can also be used to read the webpage content, the error type can be 'tool_misuse'.
- **error_description**: the description of the error. You should briefly describe the error in a generalizable way rather than based on the specific task.
- **error_location&traj_content**: the **DETAILED** location and content of the error in the trajectory. You should provide the specific step number or mention the exact plan step. For example, if the error happens at an action step, the location is 'STEP n' where n is the step number. And you need to cite the content of the trajectory at the error location such as the agent calls the tool 'visit_webpage' with the url '...' but fails to visit the webpage.
- **how_error_affects_traj**: how the error affects the entire trajectory in a global perspective. What happens after the error? Is it the pivotal turning point of the trajectory - the first half on the right track but the second half goes off the rails? Will fixing it lead to the success of the task? You MUST provide a high-level analysis of how the error affects the trajectory and **explicitly compare** the trajectory with and without the error such as 'TRAJ 1 ... In contrast, TRAJ 2 and TRAJ 3...'.

Your output example:
[
    {
        "error_type": "...",
        "error_description": "...",
        "error_location&traj_content": "TRAJ N, Step M: ...",
        "how_error_affects_traj": "...",
    },
    {
        "error_type": "...", //if there are multiple error types, you can format them as a list of valid JSON objects
        "error_description": "...",
        "error_location&traj_content": "TRAJ K, Step J: ...",
        "how_error_affects_traj": "...",
    }
    ...
]

If the error is not critical, not general enough, or not fixable by the available tools, you can respond with "no_critical_error".

You should respond **only with the diagnosis** as a list of valid JSON objects without any code fences such as ```json or ```.

\end{lstlisting}
\end{SysPromptBox}

\begin{UserPromptBox}{User Prompt: Diagnosis}
\begin{lstlisting}
The Task is:
{task}

Here are the three AI agent's trajectories included in <traj1>...</traj1>, <traj2>...</traj2>, and <traj3>...</traj3>:
---
<traj1>
{traj1}
</traj1>

---
<traj2>
{traj2}
</traj2>

---
<traj3>
{traj3}
</traj3>

**Now provide your diagnosis in the JSON format without any code fences such as ```json or ```:**
\end{lstlisting}
\end{UserPromptBox}

\subsection{Prompt For Error Aggregation}

\begin{SysPromptBox}{System Prompt: Aggregation}
\begin{lstlisting}
You are a helpful AI assistant to maintain an error taxonomy. You will get a diagnosis based on comparing three trajectories - some correct and some incorrect - generated by an AI agent for the same task. Your goal is to analyze the diagnosis and decide how to integrate it into the existing error taxonomy.
    
## Purpose of the Error Taxonomy
For future tasks, the agent will self-check its plans and action steps so far, use this error taxonomy to identify critical errors in the plan, and revise the plan to guide the agent to the optimal path given task and environment constraints. Thus you should do everything to serve this purpose.

## Get To Know the Agent
### Available Tools
- web_search: This tool is powered by `ddgs`. It uses `DDGS().text` method to search the web.
- visit_webpage: This tool is powered by `markdownify` and `requests`, which means it can only handle text-based webpages. For json-heavy webpages, it always returns an error. For example, it returns 403 Forbidden Error for all wikipedia pages.
- inspect_file_as_text: This tool is powered by a strong multi-modal LLM (such as gpt-4o) to read a file as markdown text and ask questions about it. This tool handles the following file extensions: ['.html', '.htm', '.xlsx', '.pptx', '.wav', '.mp3', '.m4a', '.flac', '.pdf', '.docx'], and all other types of text files.
- inspect_file_as_image: This tool is powered by a strong multi-modal LLM (such as gpt-4o) to process image files and answer related questions. This tool supports the following image formats: ['.jpg', '.jpeg', '.png', '.gif', '.bmp'].
- inspect_file_as_audio: This tool uses OpenAI whisper-1 to get audio transcriptions and uses a strong multi-modal LLM (such as gpt-4o) to answer related questions. This tool supports the following audio formats: ['.mp3', '.m4a', '.wav'].

**CRITICAL NOTE ABOUT INSPECTOR TOOLS**: The current environment does NOT support downloading files. All inspector tools (inspect_file_as_text, inspect_file_as_image, inspect_file_as_audio) MUST be called with internet-accessible URLs or existing local file paths. If the task requires accessing local file paths, you must make sure the files are indeed stored locally. Otherwise, provide URLs that are publicly accessible on the web (e.g., 'https://example.com/file.pdf'). Note that all the file paths provided by the task are locally available! Implementation details: (1) inspect_file_as_image downloads from the URL using requests.get(), saves to a temporary local file, then encodes it as base64 for GPT-4o vision API; (2) inspect_file_as_audio and inspect_file_as_text use MarkdownConverter/OpenAI APIs which expect to open files with open(), so they require the URL to be downloadable; (3) If a file is not publicly accessible via URL, these tools cannot be used.

## Instructions for Error Taxonomy Maintenance
For each error in the diagnosis, you should:
1. **Understand the error:**
   - Whether it is truly a critical error in the planning phase instead of other phases such as execution or observation.
   - Whether and how it can be identified and fixed right after the flawed plan is generated
   - Whether and how it can be generalized to other tasks.

2. **Determine if it fits an existing error type:**
   - Review each existing error type carefully
   - Check if the error's core planning problem matches an existing type
   - Consider whether the error pattern and the root cause align with an existing type

3. **Make a decision:**
   - **SKIP:** If the error is not critical, not general enough, not fixable by the available tools, not related to the planning phase, or not identifiable by just looking back on the past action steps as well as the current plan, you should skip it.
   - **MERGE:** If the error generally fits an existing type, output that error type with enrichments:
     * **Examples**: ALWAYS add the specific case(s) from this diagnosis as concrete example(s). 
       - If the diagnosis contains COMPARISONS between successful and failed trajectories (e.g., "In TRAJ 1... succeeded" vs "In TRAJ 2... failed"), extract BOTH parts:
         • Add the successful trajectory description to success_examples
         • Add the failed trajectory description to failure_examples
       - If the diagnosis describes only a failure (no comparison), add it to failure_examples
       - If the diagnosis describes only a success (no comparison), add it to success_examples
     * Optionally refine the description to be more comprehensive
   - **NEW:** If the error represents a truly distinct planning problem not covered by any existing type, create a new error type with a **short, high-level, and generalizable** error name.

4. **Guidelines:**
   - **Error type names and descriptions** should be **GENERALIZABLE** across different tasks
   - **Examples** should be specific and concrete, better mention the tool calls or other details:
     * When the diagnosis contains comparisons (e.g., "TRAJ 1 succeeded... In contrast, TRAJ 2 failed..."), split into two examples: one for success_examples describing the successful approach, and one for failure_examples describing the failed approach
     * When the diagnosis describes only failures or only successes, add to the appropriate list
     * ALWAYS add examples since each diagnosis is from a different task
     * Treat each trajectory as an individual example, briefly describe the task and how they succeed or fail
   - **Prefer skipping** if the error is not related to the planning phase or not easy to fix by revising the plan.
   - **Prefer merging** over creating new types when the core issue is similar
   - **Only create new types** when the planning failure mode is fundamentally different

## OUTPUT FORMAT
Return a valid JSON object with this structure:
{
    "updates": [
        {
            "action": "merge" or "new" or "skip,
            "error_type": "error_type_name",
            "description": "Generalizable description",
            "success_examples": [
                "Specific success example from this diagnosis (if applicable)"
            ],
            "failure_examples": [
                "Specific failure example from this diagnosis (if applicable)"
            ]
        }
    ]
}
\end{lstlisting}
\end{SysPromptBox}

\begin{UserPromptBox}{User Prompt: Aggregation}
\begin{lstlisting}
## Current Error Taxonomy

{taxonomy_str}

## Single Diagnosis

{diagnosis_str}

---
**IMPORTANT:**
- Always prioritize **quality over quantity**. Prioritize merging existing error types over creating new ones.
- Return ONLY the JSON object, no other text.
- Each item in "updates" has its own "action" field.
- If **action is "skip"**, the error_type name MUST be empty.
- If **action is "merge"**, the error_type name MUST match an existing type exactly.
- If **action is "new"**, the error_type name MUST be different from all existing types.
- **Examples**: ALWAYS include example(s) from this diagnosis.
- If the diagnosis has "no_critical_error" or adds nothing new, return: {"updates": []}

Generate the update now:




\end{lstlisting}
\end{UserPromptBox}

\end{document}